# Uniform Discretized Integrated Gradients: An effective attribution based method for explaining large language models

**Swarnava Sinha Roy and Ayan Kundu**
Wells Fargo International Solutions Private Limited
{swarnava.sinharoy, ayan.kundu}@wellsfargo.com

## Abstract

Integrated Gradients is a well-known technique for explaining deep learning models. It calculates feature importance scores by employing a gradient based approach computing gradients of the model output with respect to input features and accumulating them along a linear path. While this works well for continuous features spaces, it may not be the most optimal way to deal with discrete spaces like word embeddings. For interpreting LLMs (Large Language Models), there exists a need for a non-linear path where intermediate points, whose gradients are to be computed, lie close to actual words in the embedding space. In this paper, we propose a method called Uniform Discretized Integrated Gradients (UDIG) based on a new interpolation strategy where we choose a favorable non-linear path for computing attribution scores suitable for predictive language models. We evaluate our method on two types of NLP tasks- Sentiment Classification and Question Answering against three metrics viz Log odds, Comprehensiveness and Sufficiency. For sentiment classification, we have used the SST2, IMDb and Rotten Tomatoes datasets for benchmarking and for Question Answering, we have used the fine-tuned BERT model on SQuAD dataset. Our approach outperforms the existing methods in almost all the metrics.

1. Introduction

In recent times, the machine learning paradigm has seen incredible advancement in the field of large language models [1] [2], owing to their accurate predictions and natural language generating capabilities. While the complexity of machine learning models has itself increased, the demand for model interpretability has grown significantly especially in domains like finance, legal and healthcare, where having a transparent model is critical. Models such as BERT and its variants are being used in various NLP tasks like classification, Question-Answering, machine translation etc. Although some of these language models achieve state of the art performance on most benchmark datasets, explainability of their predictions is equally important if not more. Some of the industrywide accepted explainability techniques like LIME [11] and SHAP [3] are post-hoc in nature and may not capture the true features contributing to the model outcome. Gradient based approach is one way to explain the model predictions where the gradient of the model output is calculated with respect to the inputs (words or phrases in a text). One such method is the Integrated Gradients [4], an attribution technique commonly known for satisfying two axioms favorable for gradient based methods: Sensitivity and Implementation invariance. In the IG method, attribution of a word is calculated by accumulating gradients along a linear path from a baseline input (black image for image networks, zero embedding vector for textual models) to the target input. Although the IG method can be used for generating attribution scores of features in both continuous and discrete spaces, it has certain limitations for textual data as the points interpolated along this straight line are not good representations of the actual words or tokens due to the inherent discreteness of the underlying word embedding space. The interpolated points can be extremely far off from any word in the embedding space and computing their gradients may lead to sub-optimal attribution scores. To mitigate this, [5] proposed a technique called Discretized Integrated Gradients (DIG), where they have relaxed the constraint of choosing the intermediate points along the straight line between the baseline input and the target input. They select points close to words (anchors) in the vocabulary itself to justify correct attribution computations. This resulted in more faithful and better attribution score generations than IG.

However, DIG has a few limitations. It fails to ensure the uniformity and boundedness of the interpolated points. Based on the nature of the embedding space, these intermediate points can lie far away from the target or the reference word. It is also likely that they will not be uniformly spread between the reference token and the input word. This can create a highly non-linear path which could even violate the monotonic conditions required for computing the attribution scores defined in [5]. Consequently, the gradients computed along this path may yield erroneous attribution scores. We verify this in a subsequent section by comparing the delta approximation error of DIG and our method.

In this paper, we propose an improved approach called Uniform Discretized Integrated Gradients (UDIG) by creating a new interpolation strategy and using a new baseline token ("MASK"). The intermediate points are chosen uniformly along the straight line joining the baseline and the target input. An anchor word is then selected around each point. This ensures the uniformity and boundedness of the interpolated points required for more accurate attribution computation. We also assert monotonicity between these words to apply Riemann's summation for approximating the UDIG Integral. The choice of the baseline or reference token also plays a significant role here. DIG uses the "PAD" token as its baseline in the attribution formula. In a later section, we will show, how in our case, using the "MASK" token resulted in better performance metrics and word attributions.

We applied UDIG to generate attributions for two different tasks: Sentiment Classification and Question-Answering. For sentiment classification task we tested our interpolation algorithm to generate attributions for two large language models – BERT [1] and DistilBERT [6] each fine-tuned on three sentiment classification datasets - SST2 dataset [7], IMDB dataset [12] and Rotten Tomatoes dataset [13]. For Question-Answering task we applied UDIG on BERT model fine-tuned on the SQuAD dataset [14] . We evaluated the performance on three automated metrics viz. log odds, comprehensiveness, and sufficiency against the existing attribution methods.
Our contributions in this paper[1] are as follows:
- We have proposed a new interpolation algorithm for selecting points along a non-linear path from the input word to the baseline word.

---
[1] All codes are available in GitHub: GitHub - swarnava-sr/UDIG: Uniform Discretized Integrated Gradients

- We have used the <MASK> embedding (the embedding of "MASK" token) instead of <PAD> embedding (the embedding of "PAD" token) as the baseline.
- We have evaluated our proposed algorithm on two language tasks: Sentiment Classification and Question-Answering and showed that our approach outperforms existing methods.

## 2. Methodology

In this section, we provide the methodology of computing the attribution score for each word in a sentence. We also explain the strategy of interpolating points along the nonlinear path between the baseline and the target word.

### 2.1 Attribution score

Let $F: \mathbb{R}^{m \times n} \to \mathbb{R}$ be a language model, which takes a sequence of m words as input; n is the word embedding dimension.

Let $x_{ij}$ be the $j^{th}$ dimension of the $i^{th}$ word in a sentence.
The attribution score of $x_{ij}$ is given by,

$$UDIG_{ij}(x) = \int_{x_{ij}^k = x_j'}^{x_{ij}} \frac{\partial F(x_i^k)}{\partial x_{ij}} \, dx_{ij}^k \qquad (1),$$

where $x_j'$ is $j^{th}$ dimension of the baseline word and $x_i^k$ is the $k^{th}$ interpolated point along the path from the baseline to the $i^{th}$ word in the sentence.

The final attribution score of a word is obtained by summing $UDIG_{ij}(x)$ over j and dividing it with its norm.

$$UDIG_i(x) = \frac{\sum_{j=1}^{n} UDIG_{ij}(x)}{\|UDIG\|}$$

The above integral can be approximated using the Riemann Sum approximation of integrals based on the constraint that the interpolated points are monotonic.

$$UDIG_{ij}(x) = \sum_{k=1}^{K} \frac{\partial F(x_i^k)}{\partial x_{ij}} * (x_{ij}^{k+1} - x_{ij}^k) \qquad (2),$$

where $K$ is the number of steps between the baseline and the input word.
For any given *i*, let us assume WLOG $x_j' \leq x_j^k \leq x_j$ if $x_j' \leq x_j$,
$$x_j' \geq x_j^k \geq x_j \text{ otherwise, where } k \epsilon \{1, 2, \dots K\}$$

### 2.2 Interpolation algorithm

Here we provide the details of the process of selecting the intermediate words between the input word and the baseline word. First, given an initial input word embedding *w* and a baseline embedding $w_0$, we fix points along the linear path connecting $w_0$ and *w* in a uniform manner. Next, we look for words in the vicinity of every such point on the straight line. Finally, we ensure monotonicity along the path from the baseline to the input word.

**Setting path uniformity:** In this step we set equidistant points in the word embedding space that lie along the linear path between the input word embedding $w$ and the baseline word embedding $w_0$. These points will be random high dimensional floating-point values in the embedding space and may lie far away from any real word in the vocabulary. The Integrated gradients algorithm accumulates gradients along these points for computing the attribution score. The number of steps in the interpolation process is specified in this step.

**Nearest word:** Once we have the embeddings of the sampled points along the linear path, we search for the nearest word in the vocabulary. This can be achieved by considering a neighborhood around each point and choosing k-nearest words based on a similarity score defined between word embeddings. Out of these k neighbors, we select the anchor word based on one of two approaches viz. Greedy and Max-Count introduced in [5]. This is where non-linearity is introduced in the path. The Greedy approach computes the monotonic representation of every word in the neighborhood and picks that word which is closest to its monotonic form. The Max-Count approach chooses that word with the maximum number of monotonic dimensions. We explain what monotonicity exactly means in the next section. The two approaches are illustrated in Figure 1.

**Monotonicity:** This is to ensure that we can apply Riemann Summation to approximate the integral in (1). We have kept this step identical to [5]. A word embedding $'a'$ is defined to be monotonic w.r.t. $w$ and $w_0$, if all the dimensions of $a$ are individually monotonic with the corresponding dimensions of $w$ and $w_0$. The monotonic dimensions of $a$ is given by:
$M_a = \{j | \omega'_j \leq a_j \leq \omega_j\} U \{j | \omega'_j \geq a_j \geq \omega_j\}$ $j\epsilon\{1,2,...D\}$ where $D$ is the embedding dimension. The remaining non-monotonic dimensions are perturbed in order to achieve monotonicity entirely.

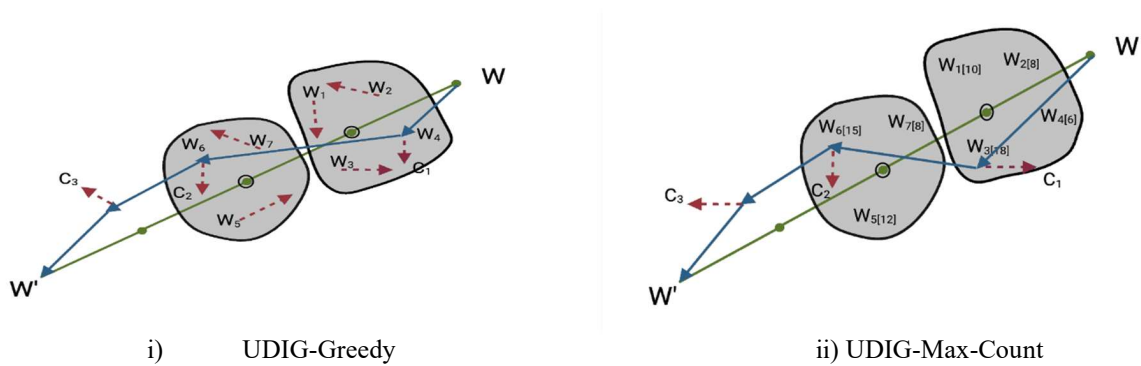

i)  UDIG-Greedy    ii) UDIG-Max-Count

Figure 1: Interpolation paths used by UDIG. W is the word of interest and W' is the baseline. The green straight line represents the linear path used by IG for calculating attribution.
(i) Grey regions are the neighborhoods of the points chosen on the straight line. Each word in this neighborhood is first monotonized where each red arrow signifies the distance between the word and its corresponding monotonic point. The word closest to its corresponding monotonic form is selected as the anchor word. (w4 since the red arrow of w4 has the smallest magnitude).
(ii) The word with the highest number of monotonic dimensions (count shown in []) is selected as the anchor word (w3 since it has the highest number of monotonic dimensions), which is followed by updating the non-monotonic dimensions to make it monotonic-c1 (red arrow). Repeating this process multiple times for each point gives the non-linear blue path for UDIG. Please refer to Section 2.2 for more details.

### 2.3 Choice of baseline

According to the authors of the Integrated Gradients paper, the baseline should be chosen such that the output of the model at the baseline is "near zero". Although in [5], authors have used the "PAD" token as the baseline, the "MASK" token satisfies this criterion better (Table 1). This could be due to the "PAD" token embedding being updated during pre-training and/or fine-tuning phase of the language model training. Whereas the "MASK" token embedding learns to fill missing words which are removed randomly from a sentence during its pre-training phase, making it equally unbiased for all the words in the vocabulary. Hence it serves as a better choice for the baseline.

| Max sequence length | Model output | |
|---|---|---|
| | <MASK> | <PAD> |
| 1 | 1.36 | 1.40 |
| 10 | 0.48 | 0.97 |
| 25 | 0.05 | 0.80 |
| 50 | 0.08 | 0.69 |
| 100 | 0.07 | 1.12 |

Table 1: Model output at baseline. "MASK" token has almost zero model output for higher sequence length

### 2.4 Axioms satisfied by UDIG

There are a couple of fundamental axioms that need to be satisfied by attribution methods as pointed out in [4]. They are    i) Sensitivity and ii) Implementation Invariance. Sensitivity axiom states that if any two inputs, that differ in only one feature, have different outcomes, then that differing feature should have non-zero attribution score. Implementation invariance axiom states that the attribution scores from two functionally equivalent models must be equal even if they differ in implementation. Our proposed method satisfies both these axioms as it is a path integral based method. Moreover, UDIG also satisfies the Completeness axiom which follows from the fundamental theorem of calculus that the sum of the attributions of an input $x$ should be equal to the difference of the model output at $x$ and the baseline $x'$ i.e., $\sum UDIG(x) = F(x) - F(x')$

| Method | DistilBERT | | | BERT | | |
|---|---|---|---|---|---|---|
| | LO ↓ | Comp ↑ | Suff ↓ | LO ↓ | Comp ↑ | Suff ↓ |
| IG | -0.950 | 0.248 | 0.275 | -0.670 | 0.237 | 0.396 |
| DIG | -1.229 | 0.301 | 0.238 | -0.835 | 0.293 | 0.371 |
| **UDIG** | **-1.653** | **0.389** | **0.165** | **-1.024** | **0.293** | **0.342** |

Table 2: Comparison of UDIG with DIG and IG on SST2 dataset

### 3. Experiment Design

In this section, we describe the datasets and the models used to evaluate our method against other explainability methods. We have followed the experimental approach used by Sanyal and Ren.

### 3.1 Sentiment Classification

**Dataset:** Our method is evaluated on three different datasets. The Stanford Sentiment Treebank (SST2) classification test dataset which consists of 1821 examples, the IMDB test dataset which consists of 25000 sentence-label pairs and the Rotten Tomatoes (RT) datasets which has 1066 labelled examples. Each dataset has binary labels for positive and negative sentiment and are available in the Huggingface library[2]. For SST2 and Rotten Tomatoes, we have used the complete test dataset, and for IMDB, we have sampled around 650 examples for our analysis.

**Language Models:** We have used the finetuned variants of the BERT and DistilBERT classification models on the SST2, IMDB and RT datasets. All the finetuned models can be downloaded from the Huggingface website.

**Baselines:** The following feature attribution methods are used for benchmarking against our method: Integrated Gradients (IG) and Discretized IG (DIG).

---

[2] Huggingface: https://huggingface.co/

**Evaluation metrics:** We have generated the same automated metrics provided in [5]; Log Odds (LO↓)[3], Comprehensiveness (Comp↑)[4] and Sufficiency (Suff↓). For LO and Comp, the top 20% most important words/tokens according to the attribution scores are masked. For Suff, the remaining 80% important words are removed from the sentences. Each time the difference in model prediction is calculated and finally averaged over all the examples. Please refer [5] for more details on how the evaluation metrics are calculated.

| Method | DistilBERT | | | BERT | | |
|---|---|---|---|---|---|---|
| | LO ↓ | Comp ↑ | Suff ↓ | LO ↓ | Comp ↑ | Suff ↓ |
| IG | -0.569 | 0.209 | 0.166 | -0.598 | 0.145 | 0.357 |
| DIG | -0.566 | 0.218 | 0.174 | -0.739 | 0.141 | 0.321 |
| **UDIG** | **-0.966** | **0.365** | **0.133** | **-1.082** | **0.191** | **0.314** |

Table 3: Comparison of UDIG with DIG and IG on IMDB dataset

| Method | DistilBERT | | | BERT | | |
|---|---|---|---|---|---|---|
| | LO ↓ | Comp ↑ | Suff ↓ | LO ↓ | Comp ↑ | Suff ↓ |
| IG | -0.424 | 0.208 | **0.190** | -0.935 | 0.250 | **0.401** |
| DIG | -0.482 | 0.245 | 0.193 | -0.912 | 0.233 | 0.416 |
| **UDIG** | **-0.531** | **0.273** | 0.191 | **-1.073** | **0.280** | 0.409 |

Table 4: Comparison of UDIG with DIG and IG on Rotten tomatoes dataset

### 3.2 Question-answering

For Question-answering task, we need two logits one for the start position of the answer and one for the end position. Based on these two logits, the span of the answer is determined. In Classification task, we have a global representation of the input from the model which is used to get the predictions. On the other hand, in Question-Answering to predict the answer span, we need to consider all the tokens to calculate the probability distributions over each word.

**Dataset:** The Question-Answering task is evaluated on the Stanford Question Answering Dataset (SQuAD). This dataset has 87599 examples in the train set and 10570 examples in the validation set. The data contains questions, contexts, and answers along with some other metadata. From the Stanford Question Answering dataset, we have sampled 950 question answer pairs based on the length of the context available for each pair.

**Language Models:** We have used BERT Question-Answering model which is fine-tuned on the SQuAD dataset. The finetuned model is available in the Huggingface library.

| Method | BERT-Start token | | BERT- End token | |
|---|---|---|---|---|
| | Suff ↓ | Comp ↑ | Suff ↓ | Comp ↑ |
| IG | 0.472 | 0.732 | 0.491 | 0.738 |
| DIG | 0.523 | 0.724 | 0.531 | 0.735 |
| **UDIG** | **0.446** | **0.742** | **0.468** | **0.751** |

Table 5: Comparison of UDIG with DIG and IG on SQuAD dataset for start and end tokens

**Baselines:** The following feature attribution methods are used for benchmarking against our method: Integrated Gradients (IG), Discretized IG (DIG).

---

[3] The arrow signifies lower is better.
[4] The arrow signifies higher is better.

**Evaluation metrics:** We have evaluated Comprehensiveness (Comp↑) and Sufficiency (Suff↓) metrics for the Question-answering prediction task. Here, we have masked the top 50% most important tokens as per the attribution scores and calculated the difference in model outputs before and after. The "SEP" token, however, is never masked in this process as it is crucial for the model to differentiate between the tokens which are part of the question and the ones related to the context in the text.

## 4. Results and Analysis

### 4.1 Performance comparison

In Tables 2, 3, 4 and 5 we have presented the comparison of IG, DIG and UDIG in terms of the metrics stated in the previous section. We have observed that UDIG outperforms other methods consistently for both Sentiment Analysis and Question Answering classification tasks.

### 4.2 Path density effect on Delta

Path integral based attribution methods satisfy the Completeness axiom as stated in Sec 2.4. This property can be used to estimate the approximation error of integral in equation (2) and is denoted by Delta % error. It is necessary that this error rate be within a threshold for accurate computation of the attribution scores. In IG, as we keep on increasing the number of steps, delta keeps on decreasing. Similarly, for UDIG, if we insert the mean of two consecutive points between all the words in the path, we observe that the delta percent error drops. This process is referred to as up-sampling in [5]. For all our experiments we have suitably chosen $f=1$ for a low delta error percentage. In tables 6, 7, 8 and 9, we have shown that UDIG has significantly lesser delta error than DIG for all the experiments. This supports our claim that DIG tends to follow non-linear paths which can generate suboptimal attribution scores.

| Method | Delta % ↓ | |
|---|---|---|
| | DistilBERT | BERT |
| DIG | 33.04% | 43.55% |
| **UDIG** | **25.22%** | **25.89%** |

Table 6: Median delta percent error on SST2 dataset

| Method | Delta % ↓ | |
|---|---|---|
| | DistilBERT | BERT |
| DIG | 30.46% | 110.36% |
| **UDIG** | **20.11%** | **15.46%** |

Table 7: Median delta percent error on IMDB dataset

| Method | Delta % ↓ | |
|---|---|---|
| | DistilBERT | BERT |
| DIG | 34.61% | 71.84% |
| **UDIG** | **31.46%** | **25.95%** |

Table 8: Median delta percent error on RT dataset

| Method | Delta % ↓ | |
|---|---|---|
| | BERT- Start token | BERT- End token |
| DIG | 36.92% | 38.39% |
| **UDIG** | **18.78%** | **19.96%** |

Table 9: Median delta percent error on SQuAD

### 4.3 Time Complexity

One of the limitations of UDIG, is that it is computationally more expensive than IG and DIG. This is due to the interpolation algorithm involving an extra step of specifying points along the straight line and then searching for the nearest actual word. It then follows a similar mechanism to DIG where it chooses the closest monotonic word in its neighborhood. More details on time complexity are covered in the Appendix.

## 5. Conclusion

In this paper, we have suggested a new method Uniform Discretized Integrated Gradients (UDIG) for computing attribution scores for textual data at a word/token level. UDIG is a hybrid of IG and DIG as it includes both their steps. It is better than IG at calculating attribution scores because it aggregates meaningful gradients of words by following a non-linear path. It is also better than DIG because it keeps the non-linearity under control by choosing words uniformly and close to the straight-line connecting the baseline and the target word. We have evaluated this approach on two types of NLP classification problems viz Sentiment Analysis and Question Answering and compared the performances with the other methods. Our method performs better consistently for both the streams of problems. It suggests that keeping the interpolation points uniform and bounded yields better results while calculating the attribution scores. We also recommend using the MASK token as the baseline when explaining models that contain this special token. For language models where the MASK token is not available e.g., GPT, the zero embedding vector or the PAD token can be a suitable alternative for the baseline. Although we validate our method on only two language tasks, gradient based approaches like UDIG can be used to derive attribution scores for any type of predictive language model. In the future, we wish to apply this model explainability technique to a wider range of NLP tasks.

# Appendix

**Visualizing attribution scores**

In figure 2, we provide visualization of explanations of the attribution methods for positive and negative sentiment sentences along with the corresponding model prediction for the sentiment classification task. In figure 3, we provide visualization of explanations of the attribution methods for start and end positions for Question-Answering task. The green highlighted words denote positive attributions and red denotes negative attributions i.e., the explanation suggests that the green highlighted words support the predicted label whereas the red ones oppose the model prediction.

Figure 2 : Visualization of explanations by IG, DIG and UDIG. Sentences 1 and 2 are trivial cases where all three methods show correct attributions as expected. In sentences 3 and 4, UDIG and DIG provide more credible explanations than IG; whereas examples 5 and 6 are cases where UDIG and IG identified the correct words, but DIG did not. In the final example, UDIG correctly identified the attributing words, but the other methods could not.

Question: Which NFL team represented the AFC at Super Bowl 50?
Predicted Answer: Denver Broncos
**Visualizations For Start Position**

Legend: ■ Negative □ Neutral ■ Positive

IG: [CLS] Which NFL team represented the AFC at Super Bowl 50 ? [SEP] Super Bowl 50 was an American football game to determine the champion of the National Football League ( NFL ) for the 2015 season . The American Football Conference ( AFC ) champion Denver Broncos defeated the National Football Conference ( NFC ) champion Carolina Panthers 24 – 10 to earn their third Super Bowl title . The game was played on February 7 , 2016 , at Levi ' s Stadium in the San Francisco Bay Area at Santa Clara , California . As this was the 50th Super Bowl , the league emphasized the " golden anniversary " with various gold - themed initiatives , as well as temporarily su ##sp ##ending the tradition of naming each Super Bowl game with Roman n ##ume ##rals ( under which the game would have been known as " Super Bowl L " ) , so that the logo could prominently feature the Arabic n ##ume ##rals 50 . [SEP]

DIG: [CLS] Which NFL team represented the AFC at Super Bowl 50 ? [SEP] Super Bowl 50 was an American football game to determine the champion of the National Football League ( NFL ) for the 2015 season . The American Football Conference ( AFC ) champion Denver Broncos defeated the National Football Conference ( NFC ) champion Carolina Panthers 24 – 10 to earn their third Super Bowl title . The game was played on February 7 , 2016 , at Levi ' s Stadium in the San Francisco Bay Area at Santa Clara , California . As this was the 50th Super Bowl , the league emphasized the " golden anniversary " with various gold - themed initiatives , as well as temporarily su ##sp ##ending the tradition of naming each Super Bowl game with Roman n ##ume ##rals ( under which the game would have been known as " Super Bowl L " ) , so that the logo could prominently feature the Arabic n ##ume ##rals 50 . [SEP]

UDIG: [CLS] Which NFL team represented the AFC at Super Bowl 50 ? [SEP] Super Bowl 50 was an American football game to determine the champion of the National Football League ( NFL ) for the 2015 season . The American Football Conference ( AFC ) champion Denver Broncos defeated the National Football Conference ( NFC ) champion Carolina Panthers 24 – 10 to earn their third Super Bowl title . The game was played on February 7 , 2016 , at Levi ' s Stadium in the San Francisco Bay Area at Santa Clara , California . As this was the 50th Super Bowl , the league emphasized the " golden anniversary " with various gold - themed initiatives , as well as temporarily su ##sp ##ending the tradition of naming each Super Bowl game with Roman n ##ume ##rals ( under which the game would have been known as " Super Bowl L " ) , so that the logo could prominently feature the Arabic n ##ume ##rals 50 . [SEP]

Question: What project structures assist the owner in integration?
Predicted Answer: design – build , partner ##ing and construction management
**Visualizations For End Position**

Legend: ■ Negative □ Neutral ■ Positive

IG: [CLS] What project structures assist the owner in integration ? [SEP] Several project structures can assist the owner in this integration , including design - build , partner ##ing and construction management . In general , each of these project structures allows the owner to integrate the services of architects , interior designers , engineers and construct ##ors throughout design and construction . In response , many companies are growing beyond traditional offerings of design or construction services alone and are placing more emphasis on establishing relationships with other necessary participants through the design - build process . [SEP]

DIG: [CLS] What project structures assist the owner in integration ? [SEP] Several project structures can assist the owner in this integration , including design - build , partner ##ing and construction management . In general , each of these project structures allows the owner to integrate the services of architects , interior designers , engineers and construct ##ors throughout design and construction . In response , many companies are growing beyond traditional offerings of design or construction services alone and are placing more emphasis on establishing relationships with other necessary participants through the design - build process . [SEP]

UDIG: [CLS] What project structures assist the owner in integration ? [SEP] Several project structures can assist the owner in this integration , including design - build , partner ##ing and construction management . In general , each of these project structures allows the owner to integrate the services of architects , interior designers , engineers and construct ##ors throughout design and construction . In response , many companies are growing beyond traditional offerings of design or construction services alone and are placing more emphasis on establishing relationships with other necessary participants through the design - build process . [SEP]

Figure 3: Visualization of explanations by IG, DIG and UDIG for predicting the start and end position of the answers. (top) This example visualizes the important tokens for predicting the start position. In this example, UDIG outperforms the other baselines by capturing the correct tokens which have higher impact on the prediction. It correctly shows that words like "Which", "NFL", "team" from the question part are important and have positive impact on the prediction. (bottom) This example visualizes the important tokens for predicting the end position. In this example, UDIG correctly captures the word like "structures" from the question part and word like "management" have high impact on the model prediction while predicting the end token.

**Time complexity**

In the below-mentioned table, we have compared the performance of IG, DIG and UDIG for different number of steps (number of interpolation points) between the input and the baseline word.

| Method | Steps | LO ↓ | Comp ↑ | Suff ↓ | Time |
|---|---|---|---|---|---|
| IG | 30 | -0.939 | 0.242 | 0.409 | T |
| DIG | 30 | -0.912 | 0.233 | 0.416 | 2xT |
| UDIG | 30 | **-1.073** | **0.280** | **0.409** | 6xT |

Table 10: Comparison of IG, DIG and UDIG for fixed number of interpolation points on Rotten Tomatoes dataset

| Method | Steps | LO ↓ | Comp ↑ | Suff ↓ | Time |
|---|---|---|---|---|---|
| IG | 60 | -0.936 | 0.242 | 0.414 | 2xT |
| DIG | 30 | -0.912 | 0.233 | 0.416 | 2xT |
| UDIG | 10 | **-1.128** | **0.285** | **0.412** | 2xT |

Table 11: Comparison of IG, DIG and UDIG for different number of interpolation points on Rotten Tomatoes dataset

We define T as the time complexity of computing IG with 30 steps where time complexity of DIG and UDIG with same numbers of steps is 2 times and 6 times that of IG respectively. It is clear from the table 10 that UDIG has more time complexity compared to IG and DIG.
But, to have fair comparison we have fixed the time complexity and experimented to test the performance for different number of steps. It is clear from the table 11 that UDIG with less steps compared to IG and DIG performs better while having the same time complexity.